\documentclass{article}

\usepackage{arxiv}

\usepackage[utf8]{inputenc} % allow utf-8 input
\usepackage[T1]{fontenc}    % use 8-bit T1 fonts
\usepackage{hyperref}       % hyperlinks
\usepackage{url}            % simple URL typesetting
\usepackage{booktabs}       % professional-quality tables
\usepackage{amsmath,amsfonts}
\usepackage{amsfonts}       % blackboard math symbols
\usepackage{nicefrac}       % compact symbols for 1/2, etc.
\usepackage{microtype}      % microtypography
\usepackage{lipsum}		% Can be removed after putting your text content
\usepackage{graphicx}
\usepackage{natbib}
\usepackage{doi}
\usepackage[dvipsnames]{xcolor}
\usepackage{enumitem}
\usepackage[figuresright]{rotating}
\usepackage{placeins}
\usepackage{siunitx}

\usepackage{fancyhdr} % Sollte durch arxiv schon geladen sein, aber sicherheitshalber
\title{Integrating a Causal Foundation Model into a Prescriptive Maintenance Framework for Optimising Production-Line OEE}

\author{ 
    Felix Saretzky\thanks{Corresponding author. Tel.: (+352) 46 66 44 5777.} \\
    Department of Engineering\\
    University of Luxembourg\\
    6, rue Richard Coudenhove-Kalergi\\ % Break
    1359 Luxemburg, Luxembourg \\
    \texttt{felix.saretzky@uni.lu} \\
    \And
    Lucas Andersen \\
    Department of Computer Science\\
    University of Luxembourg\\
    2, place de l’Université\\ % Break
    4365 Esch-sur-Alzette, Luxembourg \\
    \texttt{lucas.andersen@uni.lu} \\
    \AND
    Thomas Engel \\
    Department of Computer Science\\
    University of Luxembourg\\
    2, place de l’Université\\ % Break
    4365 Esch-sur-Alzette, Luxembourg \\
    \texttt{thomas.engel@uni.lu} \\
    \And
    Fazel Ansari \\
    Chair of Production and\\ % Break long department names
    Maintenance Management\\
    TU Wien \\ 
    Theresianumgasse 27\\ % Combine short items if needed
    1040 Vienna, Austria \\
    \texttt{fazel.ansari@tuwien.ac.at} \\
}

\begin{document}
\maketitle

\begin{abstract}
The transition to prescriptive maintenance (PsM) in manufacturing is critically constrained by a dependence on predictive models. Such purely predictive models tend to capture statistical associations in the data without identifying the underlying causal drivers of failure, which can lead to costly misdiagnoses and ineffective measures. This fundamental limitation results in a key challenge: while we can predict \textit{that} a failure may occur, we lack a systematic method to understand \textit{why} a failure occurs. This paper proposes a model based on causal machine learning to bridge this gap. Our objective is to move beyond diagnosis to active prescription by simulating and evaluating potential \textit{fixes} to optimise KPIs such as Overall Equipment Effectiveness (OEE). For this purpose, a pre-trained causal foundation model is used as a ``what-if'' simulator to estimate the effects of potential \textit{fixes}. By estimating the causal effect of each intervention on system-level KPIs, specific actions can be recommended for the production line. This can help identify plausible root causes and quantify their operational impact. The model is evaluated using semi-synthetic manufacturing data and compared with non-causal and causal baseline machine learning models. This paper provides a technical basis for a human-centred approach, allowing engineers to test potential solutions in a causal environment to make more effective operational decisions and reduce costly downtimes.
\end{abstract}

% keywords can be removed

\begin{figure*}[ht]
    \centering 
    \includegraphics[width=\textwidth]{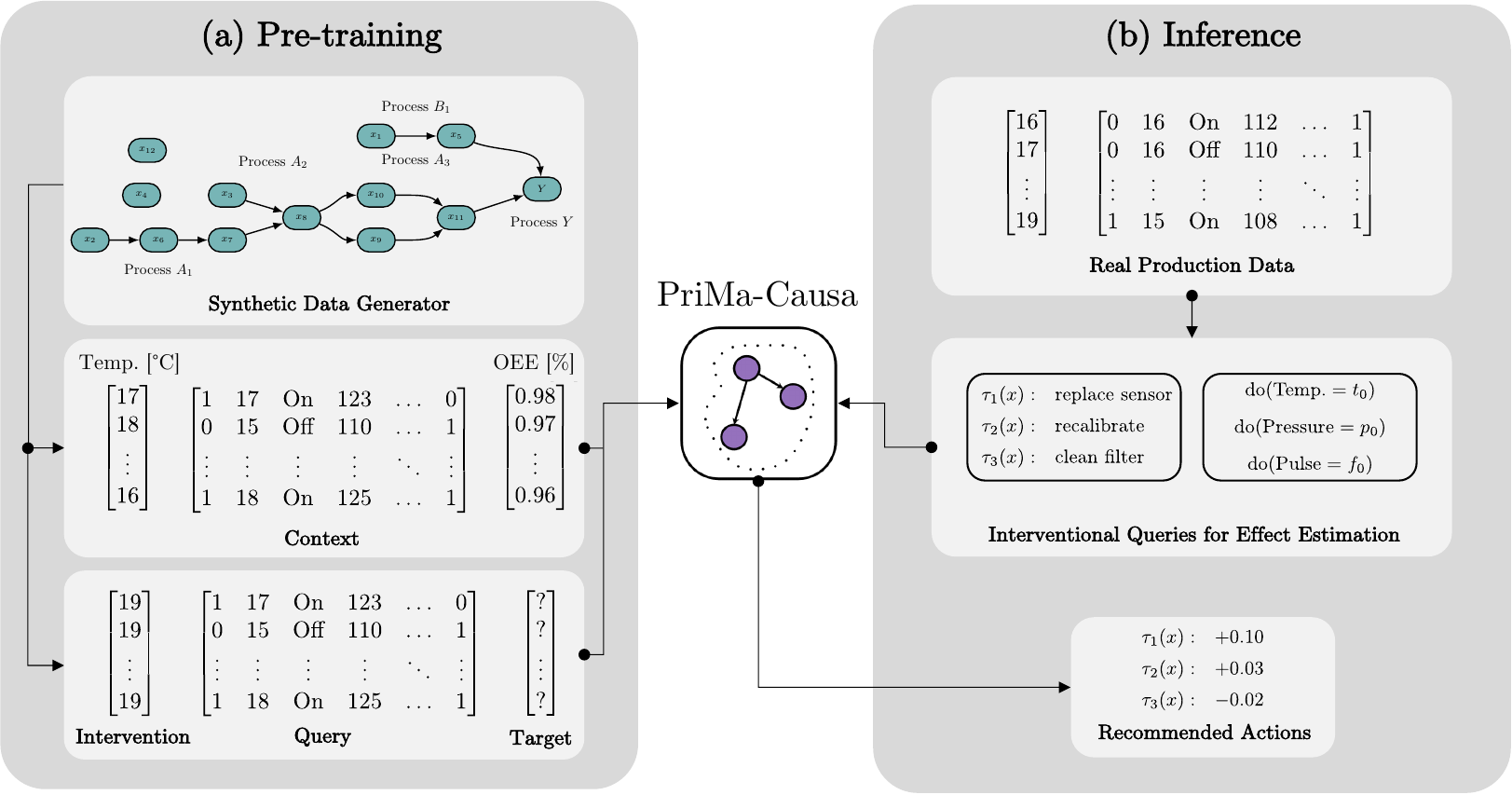}
    \caption{PriMa-Causa provides a technical basis that can extend the recommendation and decision-support layer of the PriMa framework with intervention-aware effect estimation and action ranking. (a) Pre-training (offline): A manufacturing-constrained SCM generator produces synthetic datasets reflecting sequential and physical process dependencies. From each SCM instance, an observational context $\mathcal{D}_{obs}$ and corresponding interventional targets are sampled to train a transformer-based causal foundation model via in-context learning to approximate conditional interventional distributions. (b) Inference (online): Given real production data, engineers can specify candidate maintenance actions $\tau_i(x)$ and corresponding $do$-interventions. The pre-trained model estimates $p(Y \mid do(T=t),x, \mathcal{D}_{obs})$ and the corresponding CATE values. These effect estimates are used to rank actions under practical constraints, yielding intervention recommendations.} 
    \label{fig:interventions to answer why} 
\end{figure*}
\newpage

\section{Introduction}
\label{sec: Introduction}
Prescriptive analytics (PA) is used to identify the best course of action for achieving a specific desired outcome \cite{PSM}. In contrast to descriptive or predictive models, which focus on summarising the past or predicting the future, PA aims to answer the question: ``What needs to be done/should be done, and what do I need to do to achieve this?'' \cite{PSM}. The core objective is to generate \textit{actionable decisions} \cite{Prima} that support or automate complex decision-making processes. Wissuchek and Zschech \cite{PSM} divide this decision-making process into three phases. In phase one (evaluation of alternatives), problems are identified and alternative courses of action are generated and evaluated. In phase two, the most suitable alternative is selected and executed, and phase three involves evaluating the effectiveness and results of the decision made. The extensive digitalisation of industrial processes, frequently linked to smart manufacturing and cyber-physical production systems, has generated significant research interest in data-driven decision-making processes, particularly in the context of maintenance planning as applied in PsM \cite{PSM, PsmReview2}. The objective in this context is to provide \textit{actionable recommendations} for decision-making in order to optimise future maintenance processes and minimise the probability of failures \cite{Prima,Prima-X}. The development of PsM models requires the systematic combination of information from the three fundamental stages of knowledge-based maintenance – descriptive, diagnostic and predictive analytics \cite{Prima}. Decision recommendations are generated using derived combination rules \cite{Prima} or predictions based on identified failure patterns \cite{Prima-X}. These prediction models themselves are based on statistical correlations and patterns present in the training data. To derive recommendations for action and decision-making, predictive models are often complemented by explainable artificial intelligence \cite{XAIreview}. For this purpose, post-hoc explanations \cite{Belmouadden} and intrinsic models are used to identify critical features or perform root cause analysis \cite{Muaz}. The application of these methods can lead to reliance on models whose reliability is compromised by their tendency to learn arbitrary or spurious correlations \cite{carloni}. Therefore, predictions from standard machine-learning and deep-learning models alone are often insufficient for data-driven decisions, as a comprehensive understanding of the consequences of an intervention is required \cite{prediction_to_PA}. In order to minimise the occurrence of costly misdiagnoses and ineffective technical interventions, it is essential to determine the underlying \textit{cause} of an event as well as to quantify the \textit{interventional effect}. Causal inference is a foundational methodology focused on distinguishing true cause-effect relationships from statistical associations, essential for robust decision-making and predicting the effects of deliberate interventions \cite{carloni, prediction_to_PA, pearl2009}. Causal AI enables the identification of such causal relations by modelling the underlying data-generation processes \cite{fraunhoferData}, thereby enabling reasoning across the distinct layers of the causal hierarchy: association, intervention, and counterfactual reasoning \cite{pearl2009}. Consequently, a gap exists: current PsM methods rooted in associational logic are often insufficient for estimating intervention effects and translating them into actionable recommendations. 
This paper presents three contributions:
(1) We propose PriMa-Causa, a technical basis for PsM that estimates interventional outcomes and CATEs from observational production data to rank candidate interventions under action constraints.
(2) We encode manufacturing domain knowledge via a procedural, SCM-based data generator for pre-training, constrained by sequential and physical process dependencies and supporting mixed continuous and categorical variables.
(3) We evaluate the prescriptive value of the proposed approach using semi-synthetic fast-moving consumer goods (FMCG) data with known potential outcomes. To this end, we formulate a budget-constrained intervention-prioritisation task and report expected OEE gains against causal and non-causal baselines. The paper is structured as follows. 
Section \ref{sec: Background} motivates causal intervention reasoning for PsM. Section \ref{sec: prima-c} presents PriMa-Causa, including effect estimation and the SCM-based pre-training data generator. Section \ref{sec: Use case} evaluates budget-constrained intervention ranking on semi-synthetic FMCG data. Section \ref{sec: conclusion} concludes and outlines future work.
%---------------------------------------------------------------------------------------------------------------------------------------------------------------%
%---------------------------------------------------------------------------------------------------------------------------------------------------------------%
\section{Background}
\label{sec: Background}
%---------------------------------------------------------------------------------------------------------------------------------------------------------------%
\subsection{Generating Recommendations for Action in Prescriptive Maintenance}
\label{subsec: Generating Recommendations}
According to \cite{PSM}, phase one of the three-phase decision-making process comprises problem identification, the generation of alternative courses of action, and their ranking. In this context, PriMa \cite{Prima} is a comprehensive model for PsM with a four-layer architecture. Of particular relevance to this process are layer three, the recommendation and decision support dashboard, and layer four, the semantic learning and inference layer. Layer three is responsible for aggregating the results of the predictive analyses from layer two and translating them into concrete recommendations for action, primarily through dynamic rule-based methods. These rules are triggered by combinations of quality-based parameters and the calculated remaining useful life of components. The fourth layer adds a semantic level to this process. It uses case-based reasoning to learn from past experiences and transfer problem solutions to new, similar cases. In addition, ontologies are used to formally structure maintenance knowledge. The combination of these two layers enables PriMa to not only make predictions, but also to generate optimised recommendations for action and to improve them through continuous learning from previous decisions.
In contrast, Padovano et al. \cite{Padovano} utilise the digital twin as a scenario analysis and testing module to simulate different maintenance and production schedule alternatives and compare their forecasted key performance indicators (KPIs) to determine the best-performing plan. Similarly, Torayev et al. \cite{Torayev} employ a reinforcement learning agent that, rather than simulating all alternatives, learns an optimal policy to sequentially recommend the most cost-effective manufacturing configurations, proving robust in stochastic environments.
%---------------------------------------------------------------------------------------------------------------------%
\subsection{Challenges with Standard Machine Learning and XAI in Prescriptive Maintenance}
\label{subsec: Problems with ML and XAI}
Production machines and systems are complex, interconnected technical systems in which faults can propagate across process steps, such that a single root cause can lead to different symptoms at different locations. In such environments, the causal mechanism is typically determined by physical processes \cite{saretzky}. Another challenge is the dynamic nature of the production process. Due to wear and tear, process modernisation, constantly changing customer requirements and shorter product cycles, components or systems are iteratively replaced or completely retooled. This can limit the availability of data regarding known and unknown causes of machine failures, as more data are typically available from high-OEE production than from failure events or interventions \cite{fraunhoferData, dataScienceManufacturing}. Modelling spatio-temporal dependencies of sensor data as illustrated in Figure \ref{fig:interventions to answer why} requires a methodology that goes beyond the predictive accuracy of machine learning models to decide how to intervene. Standard machine learning methods are typically trained to minimise a prediction error, for example via the mean squared error. However, as Doutreligne and Varoquaux \cite{Doutreligne} demonstrate, this approach is insufficient for causal questions, as a model with high predictive quality can systematically provide inaccurate causal estimates. A primary driver of this discrepancy is confounding bias, where unobserved common causes create spurious correlations between the intervention and the outcome \cite{carloni, prediction_to_PA, Doutreligne}. Therefore, relying on purely correlative models that fail to distinguish between spurious correlations in the data and true causal effects carries a substantial risk of producing misleading and potentially costly recommendations for action.
%---------------------------------------------------------------------------------------------------------------------------------------------------------------%
\subsection{Causal Machine Learning in Technical Processes}
\label{subsec: Causal Methods in Technical Processes}
Several studies use causal artificial intelligence frameworks to improve decision-making in industrial manufacturing and maintenance systems \cite{doubleML, fraunhoferCausalAI, causalPSM, OSEE}. The challenge is to decouple prescriptive decision-making from potentially error-prone correlations and replace it with a causal foundation. Schwarz et al. \cite{doubleML} developed a framework that uses double machine learning to estimate the causal effect of rework decisions and thus derive a profit-increasing guideline for semiconductor manufacturing. Vanderschueren et al. \cite{causalPSM} use generative models for causal inference to create individualised and cost-efficient PsM plans from service data based on the estimation of potential outcomes. Madreiter and Ansari \cite{OSEE} proposed an Overall Sustainable Equipment Effectiveness (OSEE) framework using Dynamic Bayesian Networks to model the causal interdependencies between operational, environmental, and social indicators, supporting prescriptive decision-making through scenario-based simulations. These application-specific approaches are part of a broader trend that Youssef \cite{fraunhoferCausalAI} describes as causality-driven AI, a paradigm that uses methods such as structural causal models (SCMs) to create robust solutions for process optimisation and fault diagnosis. The overarching objective of these studies is to replace purely correlative prescriptions with explicit cause-and-effect models, thereby ensuring that the decision-making frameworks are robust, unbiased, and actionable.

\subsection{Overview of Existing Research}
\label{subsec: summary of research gaps}
Based on Section \ref{sec: Introduction}, Subsections \ref{subsec: Generating Recommendations} and \ref{subsec: Problems with ML and XAI}, we summarise the identified limitations of prior research as follows:
\begin{enumerate}[label=(\roman*)]
    \item \textbf{Scalable use of domain knowledge:} PsM recommendations often rely on expert knowledge encoded as rules, ontologies, or heuristics. However, such representations are typically asset-specific and require substantial manual engineering, which limits their transferability and scalability across machines and production lines \cite{Prima,Prima-X}.
    \item \textbf{Limited interventional evidence:} Robust prescriptive maintenance is limited by the lack of structured data linking root causes, implemented maintenance actions, and observed outcomes. Moreover, collecting counterfactual evidence through randomised trials is typically impractical and costly in production settings \cite{dataScienceManufacturing}.
    \item \textbf{Prediction $\neq$ prescription:} Standard predictive and XAI-based approaches optimise correlational accuracy and feature attribution, which do not guarantee valid estimates of intervention effects under confounding, and can therefore lead to unreliable prescriptions \cite{carloni,prediction_to_PA,Doutreligne}.
    \item \textbf{Incomplete causal structure at deployment:} Many causal pipelines depend on an explicit causal graph for identification. In manufacturing, however, only partial process knowledge is often available, and full causal graphs are difficult to derive or validate at scale \cite{fraunhoferData,saretzky,OSEE}.
\end{enumerate}
%---------------------------------------------------------------------------------------------------------------------------------------------------------------%
\section{PriMa-Causa}
\label{sec: prima-c}

\subsection{Technical Basis}
\label{subsec: technical basis}

To overcome the limitations summarised in Subsection \ref{subsec: summary of research gaps}, we introduce PriMa-Causa, illustrated in Figure \ref{fig:interventions to answer why}, as a causal technical basis for actionable PsM decision support. Consistent with the evaluation-of-alternatives phase of the prescriptive-analytics decision process \cite{PSM}, at its core, PriMa-Causa's pre-trained causal foundation model, tailored to production machines, can be queried as a ``what-if'' simulator to predict target KPIs under interventions based on observational data. Since domain knowledge about manufacturing-system interactions is integrated during pre-training via the SCM-based synthetic data generator described in Subsection \ref{subsec: Synthetic Data Generator}, interventional estimates are available without retraining and without requiring a full causal graph at inference time. We propose a two-step decision logic for generating actionable recommendations:
(1) testing of plausible root-cause hypotheses to explain context-specific KPI deviations and (2) comparative evaluation of feasible maintenance strategies. Using Figure \ref{fig:interventions to answer why} as a reference production machine, the model in (1) addresses the \textit{reverse causal question}, for example \begin{color}{blue}Q1\end{color}: ``Why was the quality $Y$ classified as non-conforming for the last ten produced units?''. While a reverse causal question may not have a single, clearly defined answer \cite{why_ask_why}, we propose using a series of testable \textit{forward causal questions}, in the form of ``what-if'' simulations, to identify potential root causes. Q1 is addressed by estimating the target KPI outcome $Y$ (e.g., quality or OEE) under an intervention on a candidate root-cause variable $B_1$. Instead of the observed distribution, the conditional interventional distribution $p(Y | do(B_1 = b_{median}))$ is estimated to quantify the impact of the ``what-if'' condition. This enables reliable simulation of the causal effects of potential causes on the target metric. Thus, it is possible to validate whether an intervention at $B_1$ would remedy the observed non-conforming quality in the target variable $Y$.
In (2), PriMa-Causa can be used to answer forward causal questions. This forms the basis for prescriptive individualised decision-making. Questions such as \begin{color}{blue}Q2\end{color}: ``What is the optimal intervention $T$ to optimise a defined target metric $Y$, given the specific system context $X$?'' are addressed. The core metric for quantifying this context-specific impact is the Conditional Average Treatment Effect (CATE), formally defined as:
\begin{equation} 
\tau(x) = E[Y | do(T=1), X=x] - E[Y | do(T=0), X=x] 
\end{equation} 
CATE quantifies the expected difference in outcome $Y$ when performing an intervention $T=1$ versus a reference action $T=0$, given the specific conditions $X=x$. The estimated CATE can be used to identify interventions that are likely to increase OEE before any physical changes are made on the machine. The resulting outputs consist of a ranked set of root-cause hypotheses from (1) and context-specific CATE estimates for feasible maintenance actions from (2). Together, they support prescriptive decision-making by enabling maintenance experts to identify the most influential drivers and compare intervention alternatives before implementation, thereby answering \begin{color}{blue}Q1\end{color} and \begin{color}{blue}Q2\end{color}.
%---------------------------------------------------------------------------------------------------------------------------------------------------------------%
\subsection{Causal Effect Estimation}
\label{subsec: Causal Effect Estimation}
Following recent work on causal foundation models \cite{causalFM, doPFN}, PriMa-Causa is based on the Prior-data Fitted Network (PFN) framework \cite{PFNs, TabPFN}, a methodology employing a transformer architecture to perform Bayesian inference through in-context learning. This approach is adapted for causal effect estimation by pre-training the model exclusively on synthetic datasets generated from a diverse prior over SCMs. During inference, PriMa-Causa employs in-context learning to estimate conditional interventional distributions from observational data. Consequently, the need for dataset-specific retraining and for a causal graph is eliminated. Robertson et al. demonstrate in \cite{doPFN} that a PFN-based foundation model can predict interventional outcomes from observational data, and prove that it provides an optimal approximation of the conditional interventional distribution (CID) with respect to the chosen prior over data-generating models. Ma et al. also show that their CausalFM framework \cite{causalFM}, pre-trained with an SCM-based prior, achieves competitive results in estimating CATEs, without the need for retraining at inference time. 
%---------------------------------------------------------------------------------------------------------------------------------------------------------------%
\subsection{Synthetic Data Generator}
\label{subsec: Synthetic Data Generator}
Pre-training a PFN requires specifying a prior distribution over the underlying data-generating processes. To define this prior, we introduce a procedural SCM-based generator whose topology is constrained to reflect the known sequential and physical dependencies of production lines. The primary goal of this generator is to provide diverse and causally grounded training data for causal effect estimation \cite{causalFM, doPFN}. The generation of a single training batch follows a fixed sequence of steps to ensure causally interpretable data. Since production processes often have categorical variables (machine states, tool type, quality, etc.) as well as continuous variables (temperature, pressure, cutting speed, etc.), our data generator is based on a Mixed Additive Noise Model (MANM) \cite{merit}. 
As a first step, a directed acyclic graph is instantiated, which maps a technical process chain, as in Figure \ref{fig:interventions to answer why}. The graph is generated as a causal process graph (CPG) \cite{saretzky} using topological sorting, whereby the edge probability decreases exponentially with the node distance and the maximum input degree is limited. The graph is also constrained to have a single sink node, which represents the end of the production line. In the second step, a structural causal model is formulated. The abstract CPG is enriched with concrete functional mechanisms. Each node $X_i$ is assigned a structural equation. For continuous target variables, the MANM according to Yao et al. \cite{merit} uses an additive noise model. Here, a variable $X_i$ is modelled as a linear or non-linear function of its parent variables $PA_i$ and an additive independent noise term $N_i$
\begin{equation}
    X_i := f_i(PA_i)+N_i.
\end{equation}
This variable can describe, for example, the production yield $X_i$ of a process as a function of temperature and pressure $PA_i$ and added randomness $N_i$. For categorical variables, we use a K-level argmax model as in \cite{merit}. A latent score is calculated for each of the $K$ possible categories of a variable $X_i$. This score is composed of a function of the parent variables $PA_i$ and a noise term $N_{i,k}$ specific to this category. The variable $X_i$ ultimately assumes the category with the maximum score.
\begin{equation}
    X_i := \operatorname{argmax}_{k \in \{1,\ldots,K\}}(f_{i,k}(PA_i) + N_{i,k})
\end{equation}
This variable can describe, for example, a welding process whose result is classified into quality classes 1, 2 or 3 ($K = 3$). After assigning each node as either categorical or continuous, the SCM is fully specified. In the third step, this SCM is used to generate the datasets required for pre-training, see the pre-training stage in Figure \ref{fig:interventions to answer why}. First, an observational dataset $\mathcal{D}_{obs}$ is sampled from the SCM to serve as the context, simulating the data available in a real-world setting. Second, an interventional dataset $\mathcal{D}_{int}$ is generated by applying the $do$-operator (e.g., $do(T=t)$) \cite{pearl2009} within the same SCM instance to yield the ground-truth interventional outcomes. This interventional data serves as the query data for pre-training and contains the prediction target. 
%---------------------------------------------------------------------------------------------------------------------------------------------------------------%
\section{Industrial Use Case}
\label{sec: Use case}
To answer \begin{color}{blue}Q1\end{color} and \begin{color}{blue}Q2\end{color} from Subsection \ref{subsec: technical basis}, we evaluate PriMa-Causa as an interventional decision-support model in a realistic manufacturing setting. Causal evaluation is challenging because counterfactual outcomes cannot be observed directly for real production data. Therefore, we use a semi-synthetic setup that preserves real FMCG production covariates and adds a simulated intervention mechanism with known potential outcomes. Specifically, real production-state covariates $X$ (machine and process signals, product-related variables, and environmental conditions) are retained. A stochastic non-linear treatment-assignment model generates a binary parameter-adjustment action $T$. Potential outcomes $Y_0$ and $Y_1$ are then generated by a stochastic non-linear outcome model with heterogeneous effects. Both models are implemented as multilayer perceptrons with sampled architectures and random parameters to induce diverse non-linear response patterns. The factual outcome is selected according to the sampled action $T$. This preserves realistic state distributions while providing known counterfactuals for controlled evaluation. To align with the industrial objective of optimising OEE, we formulate a budget-constrained prescriptive task. Each production-state instance defines one decision point where the engineer can either apply a parameter adjustment for the next production cycle or keep the current settings. Since in practice adjustment capacity is limited, models should identify the most beneficial interventions.
%---------------------------------------------------------------------------------------------------------------------------------------------------------------%
\subsection{Results: Prescriptive Value and Expected OEE Gain}
\label{subsec: predict CID}
\begin{figure}[h]
    \centering 
    \includegraphics[height=0.3\textheight]{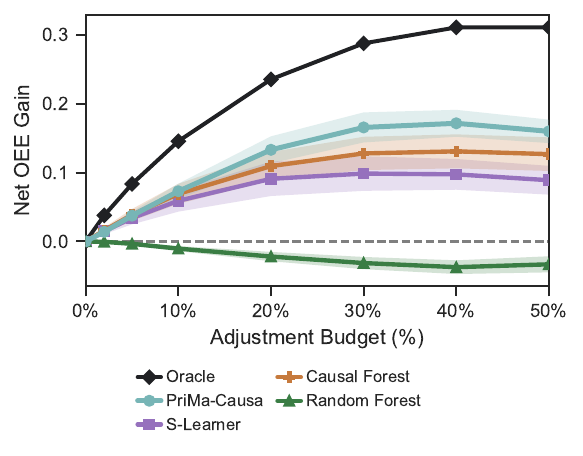}
    \caption{Expected Net OEE Gain vs. Adjustment Budget for Oracle, PriMa-Causa, Causal Forest, S-Learner, and Random Forest (non-causal baseline). Shaded bands show 95\% confidence intervals across 20 repeated runs.}
    \label{fig: OEE results} 
\end{figure}

Figure \ref{fig: OEE results} reports expected net OEE gain versus adjustment budget. For each production state, the predicted CATE is used as an intervention score, states are ranked by this score, and parameter adjustments are recommended only for the top-ranked fraction allowed by the budget, while the remaining states remain unchanged. This ranked intervention policy is then evaluated against a no-adjustment baseline using the corresponding potential outcomes, yielding expected OEE gain. We compare PriMa-Causa with an S-Learner, a Causal Forest, and a non-causal Random Forest baseline, with the Oracle reported as an upper bound. PriMa-Causa shows the strongest prioritisation in the low-to-mid budget range, while the non-causal baseline remains weaker, indicating that treatment-aware ranking improves prescriptive maintenance decisions. Gains saturate at higher budgets since the highest-impact interventions are selected first. In practice, an increased budget corresponds to longer downtime due to more interventions being tested or executed by the machine operator, which means a lower budget is preferred. 

%---------------------------------------------------------------------------------------------------------------------------------------------------------------%
\section{Conclusion and Outlook}
\label{sec: conclusion}
This paper introduced PriMa-Causa as a technical basis for prescriptive maintenance that complements correlation-based prediction with intervention reasoning. We pre-train a causal foundation model on synthetic datasets designed to reflect manufacturing processes, enabling the model to estimate interventional outcomes and CATEs from real-world production covariates $X$ and candidate interventions $T$ without additional retraining. Domain knowledge is encoded via a SCM-based generator constrained by sequential and physical dependencies typical of production lines and supporting mixed continuous and categorical variables. We evaluated PriMa-Causa in a decision-oriented setting with practical maintenance constraints. Using semi-synthetic FMCG data with known potential outcomes, we formulated a budget-constrained intervention-prioritisation task: the model ranks production states by predicted CATE and recommends adjustments under a limited action budget. PriMa-Causa achieves higher expected net OEE gains than causal and non-causal baselines, indicating that treatment-aware ranking can translate interventional estimates into actionable prescriptive policies. This supports the view that PsM decision-making benefits from intervention-aware causal modelling. Future work may expand the synthetic data generator by exploiting partial domain knowledge often available in practice (e.g., sensor ordering, process-stage constraints, or partial causal graphs) and further improve interventional-outcome accuracy. Moreover, PriMa-Causa may be combined with a root-cause analysis model to enable more targeted what-if evaluations. This would also create an opportunity to explore the real-world cost-benefit trade-off of deployment in manufacturing environments.
%---------------------------------------------------------------------------------------------------------------------------------------------------------------%
\section*{Acknowledgements}
The authors would like to acknowledge the Python libraries utilised in this research. Our implementation uses the CausalPlayground library \cite{causalplayground} to construct the SCMs. Additionally, we incorporated code from the CausalFM framework \cite{causalFM} and from TabPFN \cite{TabPFN}.

\bibliographystyle{unsrtnat}
\bibliography{reference}  %%% Uncomment this line and comment out the ``thebibliography'' section below to use the external .bib file (using bibtex) .





% \end{thebibliography}

\end{document}